\documentclass{article}
\usepackage{amssymb}
\usepackage{graphicx}
\usepackage{subcaption}
\usepackage{url}            
\usepackage{booktabs}       
\usepackage{amsfonts}       
\usepackage{nicefrac}       
\usepackage{microtype}      
\usepackage{lipsum}
\usepackage{fancyhdr}       
\usepackage{graphicx}       
\usepackage [square,numbers]{natbib}   
\graphicspath{{media/}}     
\usepackage{xspace}
\usepackage[toc,title,page]{appendix}
\newcommand{\OURS}{\textsc{EVAC}\xspace} %
\newcommand{\Ours}{\OURS} %

\newcommand{\OURSFULL}{\textsc{EnerVerse-AC}\xspace} %
\newcommand{\Oursfull}{\OURSFULL} %

\usepackage[preprint]{corl_2025} 

\newlength{\spacelength}
\title{\Oursfull: Envisioning Embodied Environments with Action Condition}

\author{
	Yuxin Jiang$^{1,*}$
	\And
	Shengcong Chen$^{1,*}$
	\And
	Siyuan Huang$^{2,*}$
	\And
	Liliang Chen$^{1,\dag}$
	\And
	Pengfei Zhou$^{1}$
	\And
	Yue Liao$^{3}$
	\And
	Xindong He$^{1}$
	\And
	Chiming Liu$^{1}$
	\And
	Hongsheng Li$^{3}$
	\And
	Maoqing Yao$^{1,\diamond}$
	\And
	Guanghui Ren$^{1, \diamond}$
}

\begin{document}
\maketitle
\renewcommand{\thefootnote}{\fnsymbol{footnote}} 
\footnotetext[1]{\hspace{-1.5mm}$^{*}$Equal contribution. $^{\dag}$Project leader. $^{\diamond}$Corresponding authors.} 
\footnotetext[2]{\hspace{-1.5mm}$^{1}$AgiBot. $^{2}$SJTU. $^{3}$MMLab-CUHK} 

\begin{abstract}
 Robotic imitation learning has advanced from solving static tasks to addressing dynamic interaction scenarios, but testing and evaluation remain costly and challenging due to the need for real-time interaction with dynamic environments. We propose \Oursfull (abbr. \Ours), an action-conditional world model that generates future visual observations based on an agent’s predicted actions, enabling realistic and controllable robotic inference. Building on prior architectures, \Ours introduces a multi-level action-conditioning mechanism and ray map encoding for dynamic multi-view image generation while expanding training data with diverse failure trajectories to improve generalization. As both a data engine and evaluator, \Ours augments human-collected trajectories into diverse datasets and generates realistic, action-conditioned video observations for policy testing, eliminating the need for physical robots or complex simulations. This approach significantly reduces costs while maintaining high fidelity in robotic manipulation evaluation. Extensive experiments validate the effectiveness of our method. Code, checkpoints, and datasets can be found at \href{https://annaj2178.github.io/EnerverseAC.github.io/}{https://annaj2178.github.io/EnerverseAC.github.io}.

\end{abstract}



\begin{figure}[h]
\centering
\includegraphics[width=1.0\textwidth]{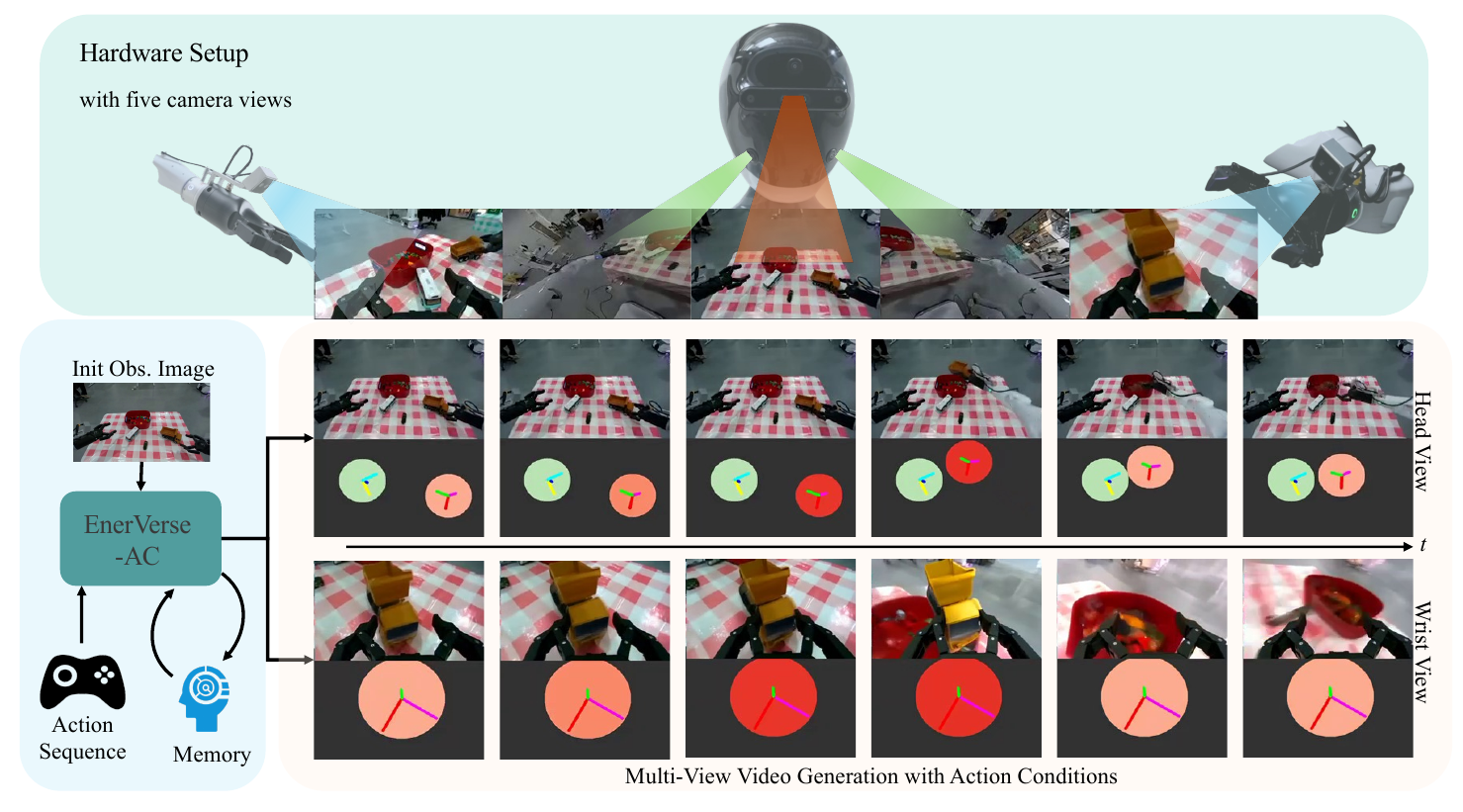}
\vspace{-0.2cm}
\caption{Overview of the \Ours framework. Given initial observation images and an action sequence, \Ours generates multi-view videos conditioned on the provided actions. By incorporating a memory mechanism, \Ours supports the generation of long-term video sequences. The framework handles both static head camera views and dynamic wrist camera views to provide a comprehensive representation of the robotic environment.}
\vspace{-0.6cm}
\label{fig:overview}
\end{figure}

\section{Introduction}

The development of robotic imitation learning has significantly advanced robotic manipulation, transitioning the field from solving isolated tasks in static environments to addressing complex and diverse interaction scenarios. Unlike traditional AI domains such as computer vision (CV) or natural language processing (NLP), where model performance can be evaluated using non-interactive and static datasets, robotic manipulation inherently requires real-time interaction between agents and dynamic environments during testing and evaluation. As task diversity grows, assessing policy performance often necessitates direct deployment on physical robots or the creation of large-scale 3D simulation environments—both of which are costly, labor-intensive, and challenging to scale.

Building low-cost, scalable testing and inference environments for robotic manipulation has thus become a critical challenge in robotic imitation learning. Recently, the concept of using video generation models as world simulators has emerged as a promising direction. These models enable agents to observe and interact with dynamic worlds through learned visual dynamics, circumventing the need for explicit physical simulation. While this approach introduces a new avenue for constructing robotic inference pipelines, existing world modeling techniques primarily focus on generating videos from language instructions and predicting actions based on the generated videos. However, these methods fall short of creating true world simulators, which should simulate environment dynamics in response to the agent’s actions, enabling realistic and controllable testing.

To bridge this gap, we propose \Ours, an action-conditional world model that generates future visual observations directly conditioned on the agent’s predicted actions. Built upon prior embodied world model architectures like \textbf{EnerVerse}~\cite{huang2025enerverseenvisioningembodiedfuture}, \Oursfull incorporates additional \textbf{A}ction-\textbf{C}onditioning information to enable more realistic and controllable robotic inference. To achieve this, we designed a multi-level action condition injection mechanism, which uses end-effector projection action maps and delta action encodings. Furthermore, to support the generation of multi-view images, crucial for embodied tasks, we introduce spatial cross-attention modules and ray direction map encoding to process multi-view features. To reflect the movement of camera, we encode the camera’s motion using ray map embeddings.

Beyond architectural innovations, the \Ours world model is designed to handle both successful and failure scenarios. In addition to leveraging the Agibot-World dataset~\cite{bu2025agibot}, we curated a diverse dataset of failure trajectories, significantly expanding the training data’s coverage. This enhancement improves the model’s ability to generalize across diverse scenarios, ensuring its applicability to real-world robotics tasks.

The proposed \Ours world model serves as both a data engine for policy learning and an evaluator for trained policy models, addressing key challenges in robotic manipulation. As a data engine, \Ours augments limited human-collected trajectories into diverse datasets by segmenting actions (e.g., fetch, grasp, home), applying spatial augmentations, and generating new video sequences, thereby enhancing policy robustness and generalization. As an evaluator, it eliminates the need for complex simulation assets by generating realistic, action-conditioned video observations for iterative policy testing, which can be reviewed by human evaluators or automated systems like Video-MLLMs. This approach significantly reduces reliance on real robot hardware during development, saving costs and time, while maintaining high evaluation fidelity correlated with real-world performance.

\section{Related Work}
\label{sec:relates}

\noindent \textbf{Video Generation Model as World Model.~} 
While prior research on generative models has shown promise, \cite{videoworldsimulators2024, bruce2024genie} highlights video generation as an innovative approach to constructing world models. Similarly, \cite{yang2024learninginteractiverealworldsimulators} aims to develop a universal world model built upon the generative model but focuses on generating only the next-step frames rather than continuous video sequences. Video generation remains a challenging task with applications across diverse domains. Recent advancements in diffusion models~\cite{DBLP:journals/corr/abs-2006-11239} and latent diffusion models~\cite{DBLP:journals/corr/abs-2112-10752} have demonstrated progress in generating high-quality images with reduced computational complexity. Furthermore, text-guided and pose-guided video generation methods~\cite{ma2024followposeposeguidedtexttovideo, hu2024animateanyoneconsistentcontrollable} have expanded the applicability of video synthesis technologies.

In robotics, works like \cite{zhou2024robodreamerlearningcompositionalworld, huang2025enerverseenvisioningembodiedfuture} focus on generating future frames from textual and visual inputs. However, limited attention has been given to video generation conditioned on robotic actions. Gesture-conditioned approaches~\cite{wang2024thisthatlanguagegesturecontrolledvideo} provide valuable insights but have yet to be tested in robotics, where environments and object interactions are significantly more complex. Advancements in action-conditioned video generation are essential to address these challenges.

\noindent \textbf{Physical Simulators for Robotics.~}  Physical simulators are widely applied in robotics learning tasks. MuJoCo~\cite{todorov2012mujoco} has been used for locomotion and manipulation studies, while PyBullet~\cite{coumans2021} supports real-time control and sim-to-real experiments. Similarly, Isaac Gym~\cite{makoviychuk2021isaacgymhighperformance} facilitates reinforcement learning in continuous control tasks with large-scale parallel environments. Several studies~\cite{DBLP:journals/corr/abs-1910-07113} utilize physical simulators to train policies for solving dexterous manipulation tasks. Despite their utility, physical simulators face notable limitations. The sim-to-real gap often results in overfitting to synthetic environments, reducing real-world performance. Moreover, creating digital assets—including robot embodiments, target objects, and task scenes—remains labor-intensive and requires expert-level effort, further hindering scalability.

\noindent \textbf{Robotics Imitation Learning.~}  Recent advancements in robotics imitation learning focus on developing generalist models capable of efficiently handling diverse tasks across multiple embodiments using extensive multimodal datasets. Models such as RT-1~\cite{brohan2023rt1roboticstransformerrealworld}, Gato~\cite{o2024open}, Octo~\cite{team2024octo}, and OpenVLA~\cite{kim2024openvla} integrate pretrained visual and language models with specialized policy heads, enabling remarkable task generalization. Building on this, \cite{bu2025synergisticgeneralizedefficientdualsystem} introduces a dual-brain system, while \cite{physicalintelligence_pi0} employs layer-wise information with flow matching techniques for action prediction. Additionally, \cite{bu2025agibot} transitions from direct action prediction to latent action representations, ensuring more effective generalization. However, these approaches rely heavily on large-scale action datasets for training. While some works, such as~\cite{huang2025adversarial}, attempt to reduce data requirements by increasing information density, they still depend significantly on human data collection, underscoring the need for further innovations in data-efficient learning techniques.




\vspace{-0.2cm}
\section{Method}
\vspace{-0.2cm}

\label{sec:method}

\begin{figure}
    \centering
    \includegraphics[width=0.95\linewidth]{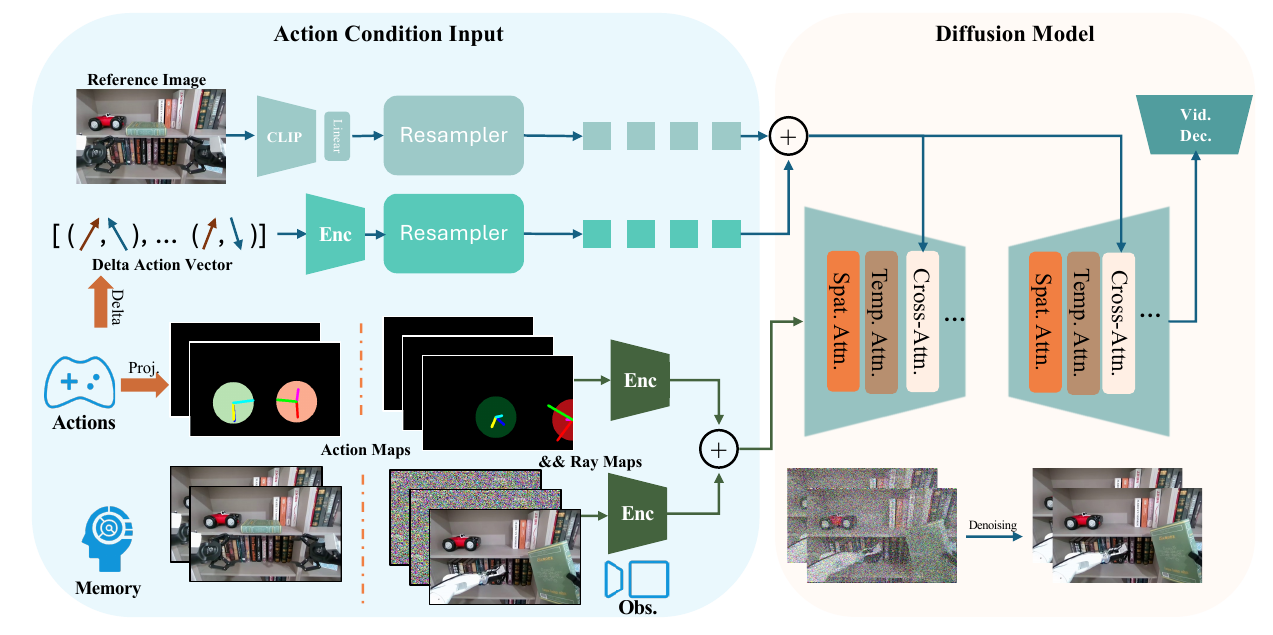}
    \vspace{-0.2cm}
    \caption{Overview of the \Ours Framework. The framework begins with a reference image, whose feature vector serves as the reference style guidance. The original robotic actions are processed to compute the delta action vector and this temporal information is concatenated with the reference style guidance and injected into the diffusion model via a cross-attention mechanism. Additionally, the action information is projected into action maps, whose feature maps are concatenated with feature maps from both memory and visual observations before being fed into the diffusion network. The diffusion model generates video frames with denoising process, followed by a video decoder to produce the final output. For simplicity, we only demonstrate the single-view case here.}
    \vspace{-0.5cm}
    
    \label{fig:framework_ace}
\end{figure}

In \Ours, we adopt a UNet-based video generation model as our baseline, following \cite{xing2023dynamicrafter, huang2025enerverseenvisioningembodiedfuture}. Beyond this, we propose an action-conditioned framework, as illustrated in Figure~\ref{fig:framework_ace}. Given a RGB video set $O \in \mathbb{R}^{V \times (H+K) \times 3 \times h \times w}$, where $V$ denotes the number of views, $H$ represents the number of observed history frames, $K$ is the number of intended predicted frames, and $h, w$ are the frame height and width, our method is designed to predict future frames based on observed past frames and robotic actions. First, we pass the video set through an encoder $\varepsilon$ to obtain the latent representation $z \in \mathbb{R}^{V \times H \times C \times h \times w }$, where $C$ is the latent dimensionality. Using a latent diffusion model, we aim to predict $z_t = p_\theta(z_{t-1},c,t)$, where $c$ is the condition signal and $t$ is the denoising timestep. In this work, the condition signal originates from the robotic action trajectory $A \in \mathbb{R}^{(H+K) \times d}$, where $d = 7$ represents the end-effector pose with $[x,y,z, roll, pitch, yaw, openness]$ and $d=14$ in bi-arm case. 


To inject the action condition, we use both spatial-aware pose information injection and delta action attention module. Furthermore, we extend traditional 2D video generation to 3D video generation, represented by multi-view frames, to better meet the requirements of robotic manipulation tasks.

\subsection{Mutli-Level Action Condition Injection}
\label{sec:mutli_level_action}




\noindent \textbf{Spatial-Aware Pose Injection.~} \cite{wang2024motionctrl}, \cite{wang2023videocomposer}, \cite{wang2024disco}, \cite{hu2024animateanyoneconsistentcontrollable} have proposed different ways on controlling video generation by injecting pose information. One common way to align the image with the fine-grained pose trajectory is to use a pixel-alignment method to inject the pose signal. In the field of robotics, end-effector 6D position has been tested as an effective representation of action space. Therefore, to ensure precise visual alignment with the conditioned image, we have developed methodologies to effectively depict the 6D end effector pose of the end effector. Firstly, we convert the end-effector position at timestamp $i$ in world coordinates to the corresponding pixel coordinates using the calibrated camera parameters. Furthermore, to visually represent the roll, pitch, and yaw angles in 2D image space, we employ visual prompting techniques inspired by~\cite{licrayonrobo, lin2024draw}. This approach utilizes unit vectors along each directional axis, providing an intuitive representation of the end-effector's orientation in 3D space. 

To illustrate the gripper action at each state, we use a unit circle to encode the action magnitude, where lighter shades correspond to open gripper and darker shades indicate closed gripper. To differentiate between the left and right hand, we employ distinct color schemes for visualizing 6D poses and gripper actions. The 6D pose visualization is rendered on a black background to enhance clarity, as shown in Figure~\ref{fig:Wrist_Head_Ray_Map}. After constructing the action map using the aforementioned visual prompting techniques, we process it with the CLIP~\cite{DBLP:journals/corr/abs-2103-00020} vision encoder. The resulting feature maps are concatenated with the feature maps from RGB images along the channel dimension.


\noindent \textbf{Delta Action Attention Module.~}  Furthermore, we designed a Delta Action Attention module which calculates the delta motion between consecutive frames to approximate changes in the end-effector's position and orientation. These delta motions are encoded into a fixed number of latent representations by a linear projector and then via cross-attention \cite{alayrac2022flamingo}, \cite{jaegle2021perceiver}. The fixed-length latent representation token is then fused with the Reference Image map and injected into the Unet stage through a cross-attention mechanism. By incorporating temporal changes, such as speed and acceleration, the module enhances the model’s physical understanding of motion dynamics, enabling it to produce more realistic and diverse video outputs.

\subsection{Multi-View Condition Injection}
\begin{figure}
    \centering
    \includegraphics[width=0.8\linewidth]{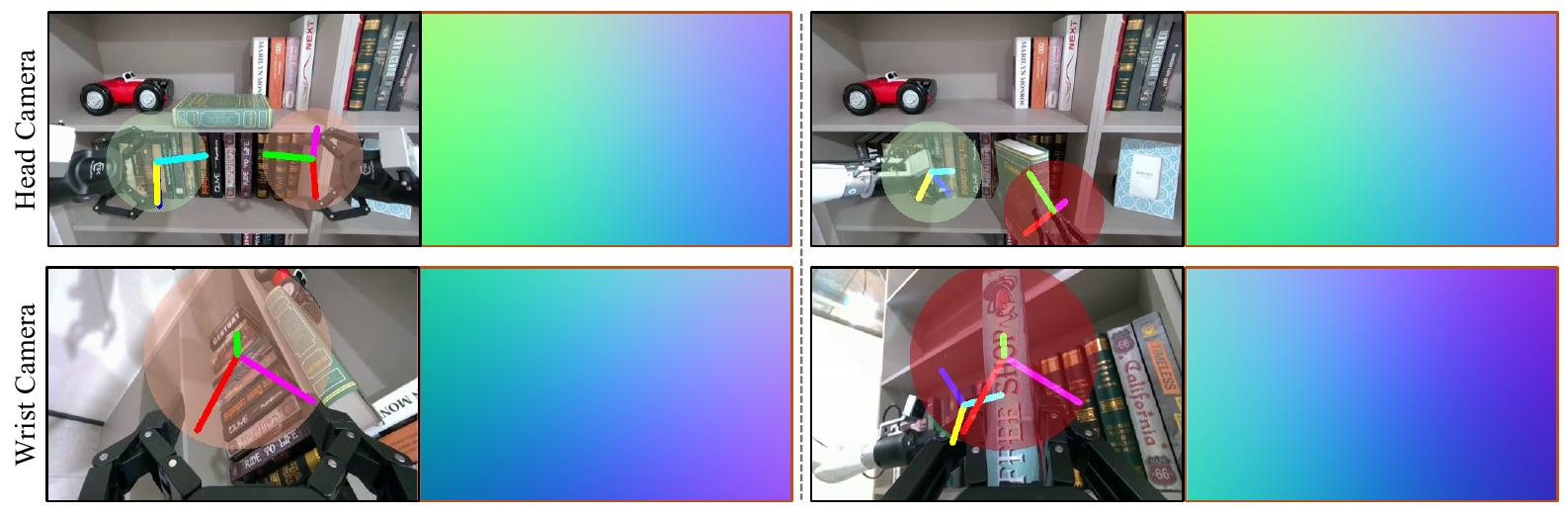}
    \vspace{-0.2cm}
    \caption{ Visualizing EEF Projections and the Ray Maps. The bottom row illustrates wrist camera views, where projections appear nearly identical. Then, ray maps provide additional spatial context to represent movements. The value of the ray maps is visualized with the RGB value.}
    \vspace{-0.6cm}
    \label{fig:Wrist_Head_Ray_Map}
\end{figure}

In embodied robotics, cross-view information, particularly visual inputs from wrist cameras, is essential for accurate trajectory prediction. To address this, we extend \Ours world model to support multi-view video generation. Following EnerVerse~\cite{huang2025enerverseenvisioningembodiedfuture}, multi-view features are fed where spatial cross-attention modules enable interaction between views. A ray direction map encoding camera parameters is also concatenated into the input features to provide spatial context. Unlike EnerVerse, which only processed static views, \Ours incorporates dynamic wrist camera views that move with the robotic arms. This creates a challenge: when projecting end-effector (EEF) poses onto wrist camera images using Section~\ref{sec:mutli_level_action} methods, the projection circle remains static, failing to convey the hand’s movement, as shown in Figure~\ref{fig:Wrist_Head_Ray_Map}.

Inspired by techniques in~\cite{gao2024cat3d,huang2025enerverseenvisioningembodiedfuture}, we encode camera motion using the origins $o_r$ and directions $d_r$ of ray maps $r=(o_r, d_r)$. Specifically, for each camera, we compute the ray maps relative to its poses at all times. Since the wrist cameras move with the arms, the ray maps of the wrist cameras can implicitly encode the motion information of EEF poses. Therefore, the ray maps are concatenated with the trajectory maps to provide enriched trajectory information, improving cross-view consistency.

\subsection{Applications}
\begin{figure}[h]
    \centering
    \includegraphics[width=0.7\linewidth]{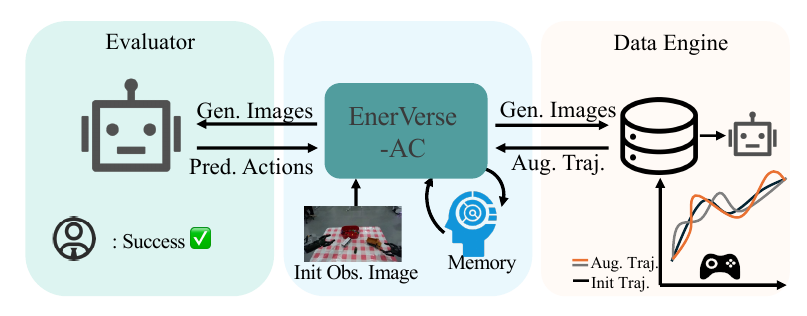}
    \vspace{-0.2cm}
    \caption{\Ours's as Data Engine and Policy Evaluator.}
    \vspace{-0.3cm}
    \label{fig:application}
\end{figure}

\noindent \textbf{Data Engine for Policy Learning.}  The \Ours world model can serve as a data engine for robotic policy learning. Specifically, for a new manipulation task (for simplicity, we use the primitive single-object pick-and-place task as an example), a human data collector first captures $M$ trajectories. For each collected trajectory, the beginning ($t_b$) and ending ($t_e$) timestamps of the gripper-object contact phase are identified by analyzing changes in gripper openness. These timestamps are then used to segment the trajectory into three distinct phases: fetching, grasping, and homing.

Focusing on the fetching phase as an example, we extract the visual observation $O_{t_b}$ and the corresponding action sequence $[a_{t_{b-N}}, \dots, a_{t_b}]$. While the action $a_{t_b}$ is kept fixed, the earlier action $a_{t_{b-N}}$ is spatially augmented to generate a new action $a'_{t_{b-N}}$. After augmentation, interpolation is applied to create new action trajectories for the sequence. Then, $O_{t_b}$ and the reversed action sequence $[a_{t_b}, \dots, a'_{t_{b-N}}]$ are fed into the \Ours world model to generate the corresponding video frames. Once the frames are generated, they are re-ordered to create a correctly sequenced dataset.  By following this process, the original $M$ trajectories can be augmented into a significantly more diverse set of trajectories, enhancing the robustness and generalization of the policy learning process.

\noindent \textbf{Evaluator for Policy Model.~} Another application of \Ours is serving as a physical simulator to evaluate trained policy models. Given an initial visual observation $O_t$ and corresponding instructions, the policy model generates action chunks. These action chunks, along with $O_t$, are then fed into the \Ours world model to generate new observations. This process is iteratively repeated until the actions generated by the policy model fall below a predefined threshold. Subsequently, multiple human evaluators watch the \Ours-generated videos to assess task success.

This evaluation approach offers two key advantages. First, it eliminates the need to create complex simulation assets, as \Ours can better represent certain physical aspects, such as fluid dynamics, compared to conventional simulators. Second, the video replay can be sped up to save time, or it can potentially be integrated with video-based Video-MLLMs, reducing the need for human evaluation efforts. By leveraging this process, the \Ours world model can largely replace the use of real robot hardware during the initial development stage, significantly reducing deployment efforts. Our experiments reveal a high correlation between evaluation results obtained through \Ours and those observed in real-world scenarios.

\section{Experiments}
\label{sec:experiment}
\vspace{-0.3cm}
\subsection{Experiment Details}
\vspace{-0.1cm}

\label{sec:exp_details}
\noindent \textbf{Dataset}
The training data for \Ours is primarily sourced from the AgiBot World dataset~\cite{bu2025agibot}, which contains over 210 tasks and 1 million trajectories. To ensure comprehensive coverage of action trajectories, including both successful and failed cases—critical for enabling \Ours to function as a generalized simulator—we collaborated with the AgiBot-Data team to gain full access to the raw data. From this dataset, we mined a substantial amount of failure cases. Additionally, we developed an automated data collection pipeline to capture real-world failure cases during teleoperation and real-robot inference, further enriching the dataset with diverse scenarios.

\noindent \textbf{Implementation Details} Our model is built on UNet-based Video Diffusion Models (VDM)~\cite{xing2023dynamicrafter}. During training, the CLIP visual encoder and VAE encoder are frozen, while other components, including the UNet, resampler, and linear layers, are fine-tuned. The model is trained with a batch size of 16. For the single-view version, training requires approximately 32 A100 GPUs for 2 days, whereas the multi-view version takes about 32 A100 GPUs for 8 days. We experimentally determined that setting the memory size to 4 and the chunk size to 16 achieves a balance between generation quality and resource cost. The memory consists of 4 historical frames, each derived from the results of the previous chunk generation. For the robotic policy model, we utilize the official single-view version of GO-1~\cite{bu2025agibot}.

\vspace{-0.3cm}
\subsection{Controllable Manipulation Video Generation} 
\vspace{-0.3cm}

\begin{figure}[h]
\centering
\includegraphics[width=0.95\textwidth]{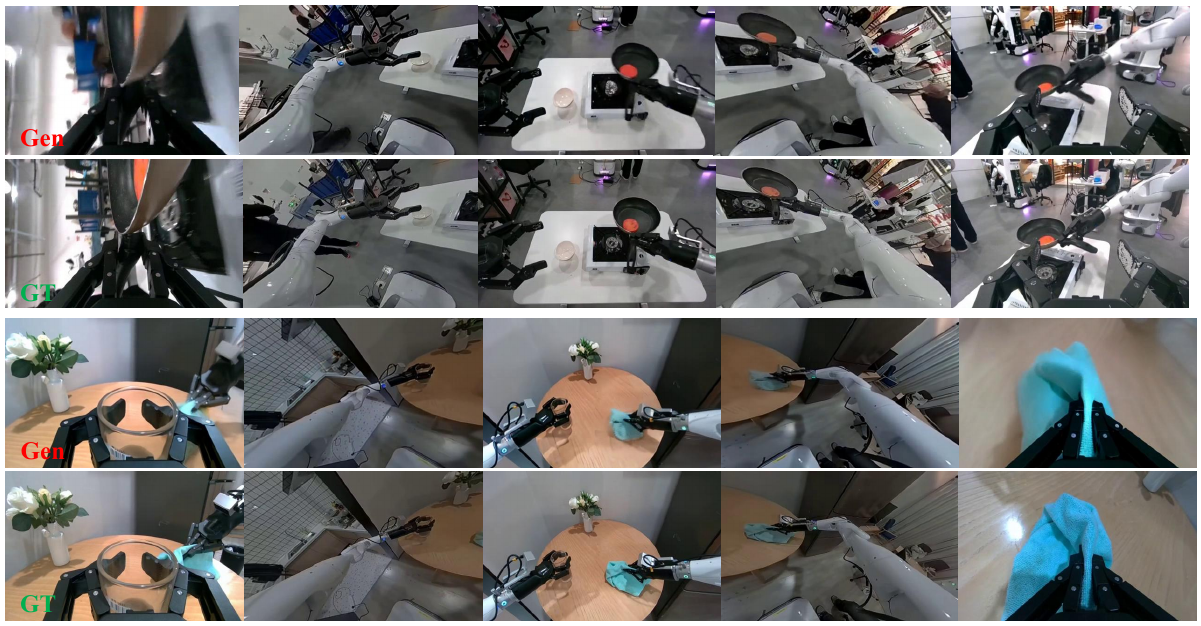}
\vspace{-0.1cm}

\caption{Qualitative results for multi-view video generation.}
\vspace{-0.3cm}

\label{quality}
\end{figure}

As shown in Figure~\ref{quality}, \Ours excels at synthesizing realistic videos of complex robot-object interactions, even in challenging scenarios. A key strength of \Ours lies in its ability to maintain high visual fidelity while accurately following input action trajectories, ensuring reliability for building credible evaluation systems.

\begin{figure}[h]
\centering
\includegraphics[width=0.95\textwidth]{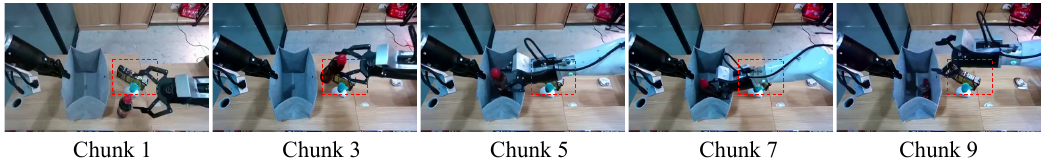}
\vspace{-0.1cm}

\caption{ Environment consistency during chunk-wise inference. Snapshots from \Ours at various inference stages (Chunks 1, 3, 5, 7, and 9) demonstrate robust performance in maintaining visual fidelity and scene coherence over time.}
\vspace{-0.2cm}
\label{env_consist}
\end{figure}

The model’s chunk-wise autoregressive diffusion architecture and sparse memory mechanism, inspired by ~\cite{huang2025enerverseenvisioningembodiedfuture}, enable it to sustain visual stability and scene consistency during continuous chunk-wise inference. Experimental results show that the generated videos remain sharp and reliable for up to 30 consecutive chunks in single-view scenarios and 10 chunks in multi-view settings. However, artifacts and blurring begin to emerge in longer sequences, highlighting a tradeoff between sequence length and visual quality. Figure~\ref{env_consist} further illustrates \Ours's ability to preserve scene integrity across multiple chunks during a manipulation task. The snapshots showcase environment consistency over time, demonstrating \Ours's robust performance in maintaining visual coherence during chunk-wise autoregressive inference.

\vspace{-0.3cm}
\subsection{\Ours as Policy Evaluator}
\vspace{-0.2cm}

\begin{figure}
    \centering
    \includegraphics[width=0.8\linewidth]{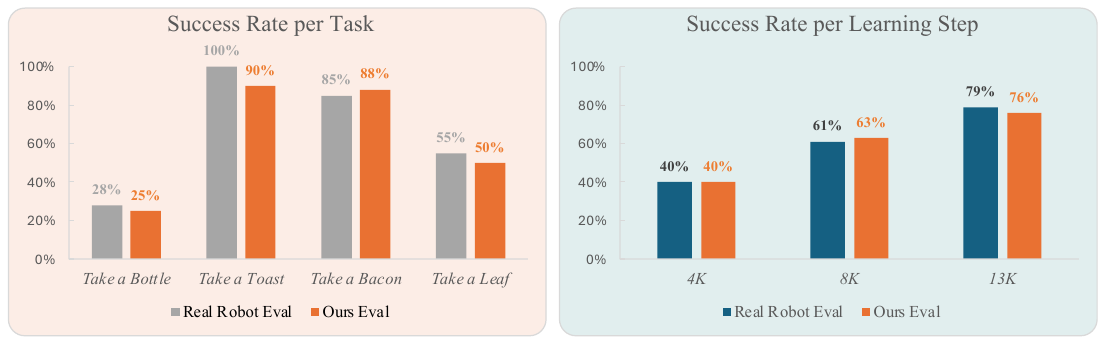}
    \vspace{-0.1cm}
\caption{Comparison of Success Rates Across Tasks and Training Steps.  
(Left)  Despite tasks vary a lot, the \Ours simulator consistently aligned its evaluation results with real-world ones.  
(Right) Success rates of a single policy model evaluated at three training steps. Both  \Ours  and real-world testing demonstrated a similar performance gradient.}
    \vspace{-0.5cm}

    \label{fig:SR_Compare_real_gen}
\end{figure}

This section evaluates the consistency between the \Ours-based generative simulator and real-world environments. Four manipulation tasks were selected for evaluation; details for these tasks can be found in the Appendix. For each task, real-world evaluations were conducted first, and the initial frame recordings from these tests were used as the image condition for \Ours evaluations. Success or failure was determined by three independent evaluators who observed either real-world executions or \Ours-generated sequences. As shown in Figure~\ref{fig:SR_Compare_real_gen} (left), while there were minor differences in absolute success rates between \Ours and real-world evaluations, the relative performance trends across tasks were consistent. These findings demonstrate \Ours's reliability for cross-task policy performance analysis and its ability to closely replicate real-world dynamics. We also provide the qualitative evaluation results in LIBERO~\cite{liu2023libero} simulator in the Appendix.


Another challenge in robot policy learning is the instability during training, e.g. performance fluctuates across training steps. To assess \Ours's ability to reflect these fluctuations, we evaluated the same policy at different training steps using the "Take a Bottle" task as an example. As shown in Figure~\ref{fig:SR_Compare_real_gen} (right), both \Ours and real-world evaluations captured the same performance trend, with success rates improving as the number of training steps increased. This result confirms that \Ours accurately mirrors real-world performance variations during policy training.

\vspace{-0.3cm}
\subsection{\Ours as Data Engine}
\vspace{-0.3cm}

In this part, we aim to demonstrate the potential of \Ours to generate novel action trajectories that augment policy training data, leading to improved task performance. The evaluation task involves picking a bottle of water from a paper box and placing it on a table. This task is challenging due to the precise force and manipulation required to extract a tightly packed bottle.

We compare two training setups: (1)Baseline: The policy is trained with only 20 expert demonstration episodes. (2)Augmented Dataset: The policy is trained with the same 20 expert episodes, augmented with 30\% additional trajectories generated using the \Ours world model.
As shown in Table~\ref{tab:data_augmentation}, the success rate (SR) improves significantly from 0.28 to 0.36 when the augmented trajectories are included in the training data. This result highlights the capability of the \Ours world model to enhance policy learning by providing diverse and effective training samples, even when the number of expert demonstrations is limited.

\begin{table}[h!]
\centering
\caption{Impact of Data Augmentation on Policy Training Success Rate.}
\label{tab:data_augmentation}
\begin{tabular}{lcc}
\hline
\textbf{Training Data}                        & \textbf{Success Rate (SR)} & \\ \hline
Baseline                       & 0.28                        & \\ 
Augmented Dataset(Adding 30\% synthetic data) & 0.36                        & \\ \hline
\end{tabular}
\vspace{-0.3cm}

\end{table}



\subsection{Further Analysis}

\begin{figure}[h]
\centering
\includegraphics[width=0.95\textwidth]{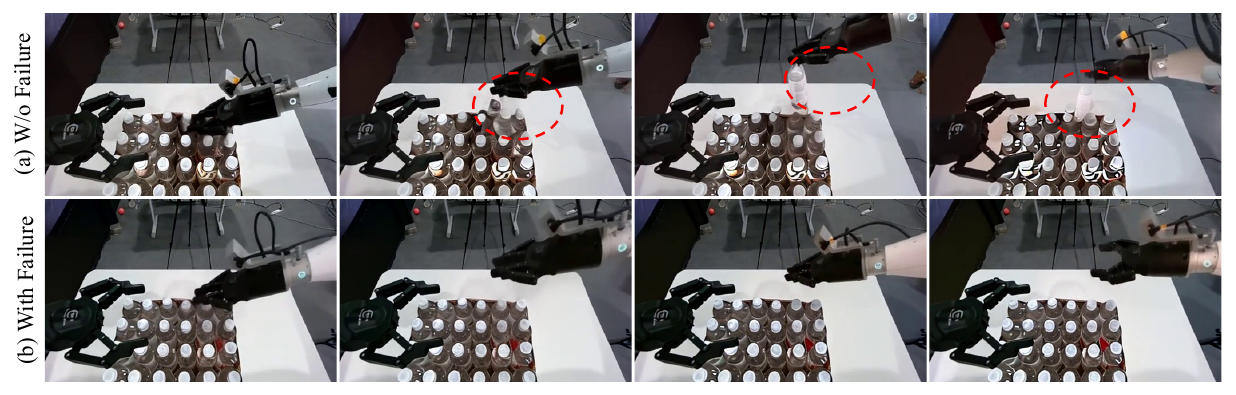}
\caption{Impact of Failure Data and Data Augmentation on Trajectory Generation. Without failure data: The model overfits to success-only trajectories, incorrectly "hallucinating" that the bottle has been grasped by the robotic arm.}
\label{fig:hallucination_wo_failure_data}
\end{figure}

\noindent \textbf{Failure Data Matters.} As discussed in Section~\ref{sec:exp_details}, we deliberately collected failure trajectories to expand the action coverage in the training data. To evaluate the effectiveness of this failure data, we trained two models: one with failure trajectories included and the other without. As illustrated in Figure~\ref{fig:hallucination_wo_failure_data}, we tested the models using a scenario where the robotic arm was pretending to grasp a bottle of water that was not actually present.

Without failure data, the model tended to overfit to successful examples, leading it to "hallucinate" that the bottle had been successfully grasped despite the absence of physical interaction. In contrast, with the inclusion of failure data, \Ours was able to accurately recognize and distinguish the failed grasp attempt, demonstrating its robustness against overfitting and its ability to handle edge cases effectively.

\section{Conclusion}
\label{sec:conclusion}
    
In this paper, we introduced \Ours, an embodied world model with action-conditioned capabilities. We proposed multi-level action condition injection strategies and utilized camera ray maps to model dynamic camera's motion. Through extensive experiments, we demonstrated the dual functionality of \Ours: serving as both a data engine for policy learning and an evaluator for trained policy models.

\clearpage

\section{Limitations and Discussions}
We proposed \Oursfull, an action-conditional embodied world model generator by implementing a video generation framework. However, several limitations remain in the current work.

First, our approach for interpreting gripper openness using a unit circle representation with varying color intensities may not generalize effectively to more complex end-effectors, such as dexterous hands. Adapting the framework to different robotic hardware configurations will require additional preprocessing steps and refinements.

Second, the wrist camera often captures irrelevant background noise, such as people moving around the robot workspace, which increases the complexity of video generation. As shown in our experiments, this limitation restricts multi-view inference to 10 chunks, compared to the 30 chunks achievable in single-view generation, thereby reducing the overall efficiency of multi-view scenarios.

Additionally, several potential applications of our action-conditioned world model remain unexplored, such as its integration with actor-critic methods for reinforcement learning. Future research could extend our framework by exploring these applications and drawing inspiration from prior work~\cite{ha2018worldmodels, hafner2022mastering, hafner2023mastering}. We hope that this work serves as a foundation for advancing embodied world models and inspires further developments in the field.



\bibliography{reference}  

\clearpage

\appendix
\section{Appendix}
In this appendix, we first provide additional training details and model parameters in Section~\ref{sec:training_details}. Next, we present an ablation study to evaluate the effectiveness of the proposed delta action attention module in Section~\ref{sec:supp_ablations}. We also include more quantitative results from both real-world and simulator experiments. Finally, we demonstrate the application setup of \Ours as a policy evaluator and a data engine in Section~\ref{sec:supp_application_setup}. For further details and to access videos with clearer indications, we strongly recommend visiting the \hyperlink{https://annaj2178.github.io/EnerverseAC.github.io/}{https://annaj2178.github.io/EnerverseAC.github.io/}.

\subsection{Training Details}
\label{sec:training_details}

As mentioned in Section~\ref{sec:method}, we adopt a UNet-based denoising model in \Ours, which takes both latent images and concatenated conditions as inputs. The concatenated conditions include repeated latent features of the condition frame, action maps, ray maps, and a dropout mask indicating whether the condition is dropped. This dropout strategy is designed to improve the model's robustness. The hyperparameters of the model architecture and training setup are provided in Table~\ref{tab:appendix_train_exp}.

\begin{table}[h!]
\centering

\begin{tabular}{l|l}
\hline
\textbf{Category} & \textbf{Configuration} \\
\hline
\hline

Diffusion Parameters & \\
- Diffusion steps & 1000 \\
- Noise schedule & Linear \\
- $\beta_0$ & 0.00085 \\
- $\beta_{T}$ & 0.0120 \\

\hline
\hline
UNet & \\
- Input channels & 19 \\
\quad - Latent image channels & 4 \\
\quad - Condition latent image channels & 4 \\
\quad - Action map channels & 4 \\
\quad - Ray map channels & 6 \\
\quad - Dropout mask channel & 1 \\
- $z$-shape & $40\times64\times4$ \\
- Base channels & 320 \\
- Attention resolutions & 1, 2, 4 \\
- Channel multipliers & 1, 2, 4, 4 \\
- Blocks per resolution & 2 \\
- Context dimension & 1024 \\
\hline
\hline

Data & \\
- Video resolution & $320\times512$ \\
- Chunk size & 16 \\
- Views & head, head\_left(\textbf{F}),  head\_right(\textbf{F}), left\_hand, right\_hand \\
\hline
\hline

Training & \\
- Learning Rate & $5\times 10^{-5}$ \\
- Optimizer & Adam($\beta_0$=0.9, $\beta_1$=0.999) \\
- Batch size per GPU (single-view) & 8 \\
- Batch size per GPU (multi-view) & 1 \\
- Parameterization & v-prediction \\
- Max steps & 100,000 \\
- Gradient clipping & 0.5 (norm) \\ 
\hline

\end{tabular}
\vspace{0.5cm}
\caption{Model Parameters and Training Configuration. \textbf{F} indicates the fisheye camera.}
\label{tab:appendix_train_exp}
\end{table}

\clearpage

\subsection{More Ablations}
\label{sec:supp_ablations}

\begin{figure}[h]
\vspace{-1cm}
\includegraphics[width=1\textwidth]{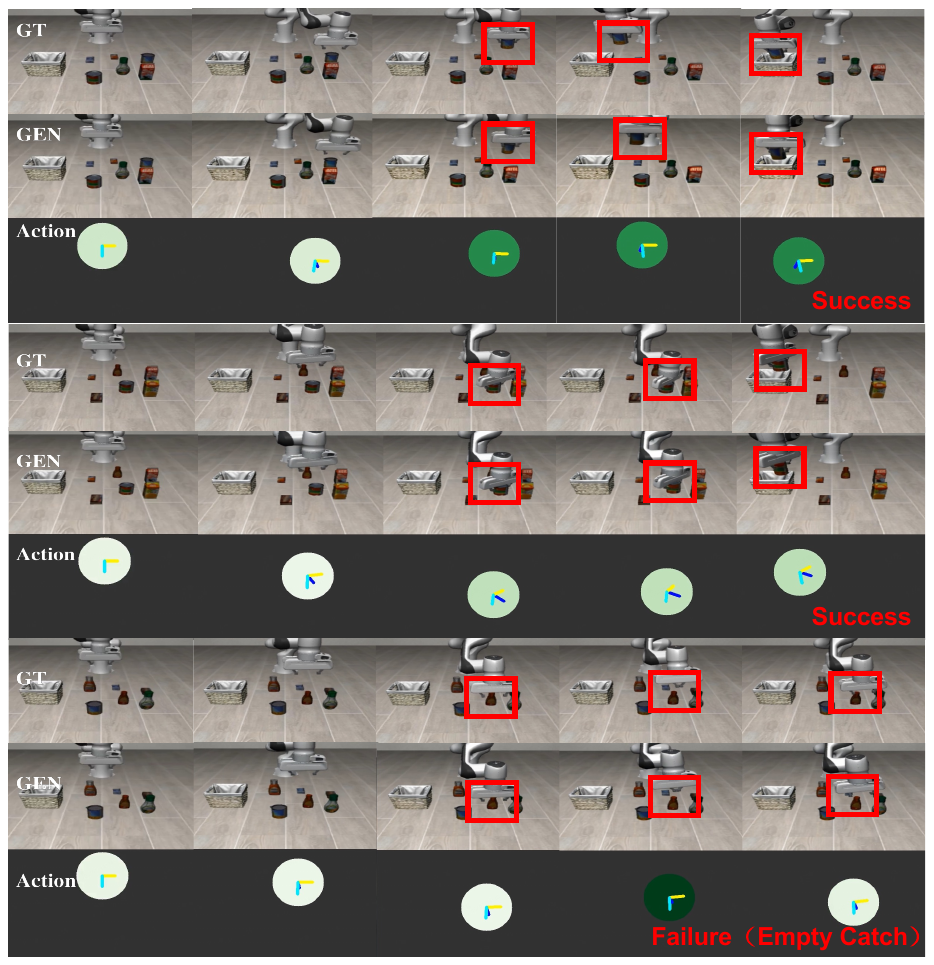}
\vspace{-2cm}
\caption{Results of generated videos under identical conditions with and without the Delta Action Module. The top row shows results with the module (w/ Delta Attention), while the bottom row shows results without it (w/o Delta Attention). The dashed red boxes highlight regions with inconsistent or hallucinated results. }

\label{fig:attention_img}
\end{figure}

\noindent \textbf{The effectiveness of Delta Action Attention Module}
To ensure the generated videos from \Ours accurately follow action trajectories, we employ a Multi-Level Action Condition Injection strategy. To assess the impact of the Delta Action Attention Module, we conducted ablation studies, with results shown in Figure~\ref{fig:attention_img}. The task involved intricate pan manipulation dynamics, including rapidly shaking the pan, slowly shaking the pan, upward tossing, and upward shaking.

The primary challenge lies in distinguishing between upward tossing and upward shaking, as they exhibit fundamentally different acceleration profiles. Upward tossing involves a sharp, high-acceleration movement, whereas upward shaking follows a smoother, low-acceleration trajectory. Without the Delta Action Module, spatial-aware action recognition models often fail to differentiate between these motions, resulting in incorrect predictions. This leads to temporal inconsistency, such as flickering or the sudden disappearance of objects (e.g., the ham) due to erratic motion transitions, as highlighted by the dashed red boxes in Figure~\ref{fig:attention_img}.

The Delta Action Module addresses these limitations by introducing acceleration-aware action decomposition. By explicitly modeling the time-derivative of actions, the module captures second-order dynamics (velocity changes), enabling it to differentiate between high-acceleration motions like tossing and low-acceleration motions like shaking. As a result, the Delta Action Module ensures significantly stronger motion consistency and reduces temporal errors compared to configurations without it.

\clearpage

\subsection{Additional Results}

\subsubsection{More Real-World Generated Multi-View Results}

\begin{figure}[h] 
    \centering
    \includegraphics[width=1\textwidth,height=1.3\textwidth]{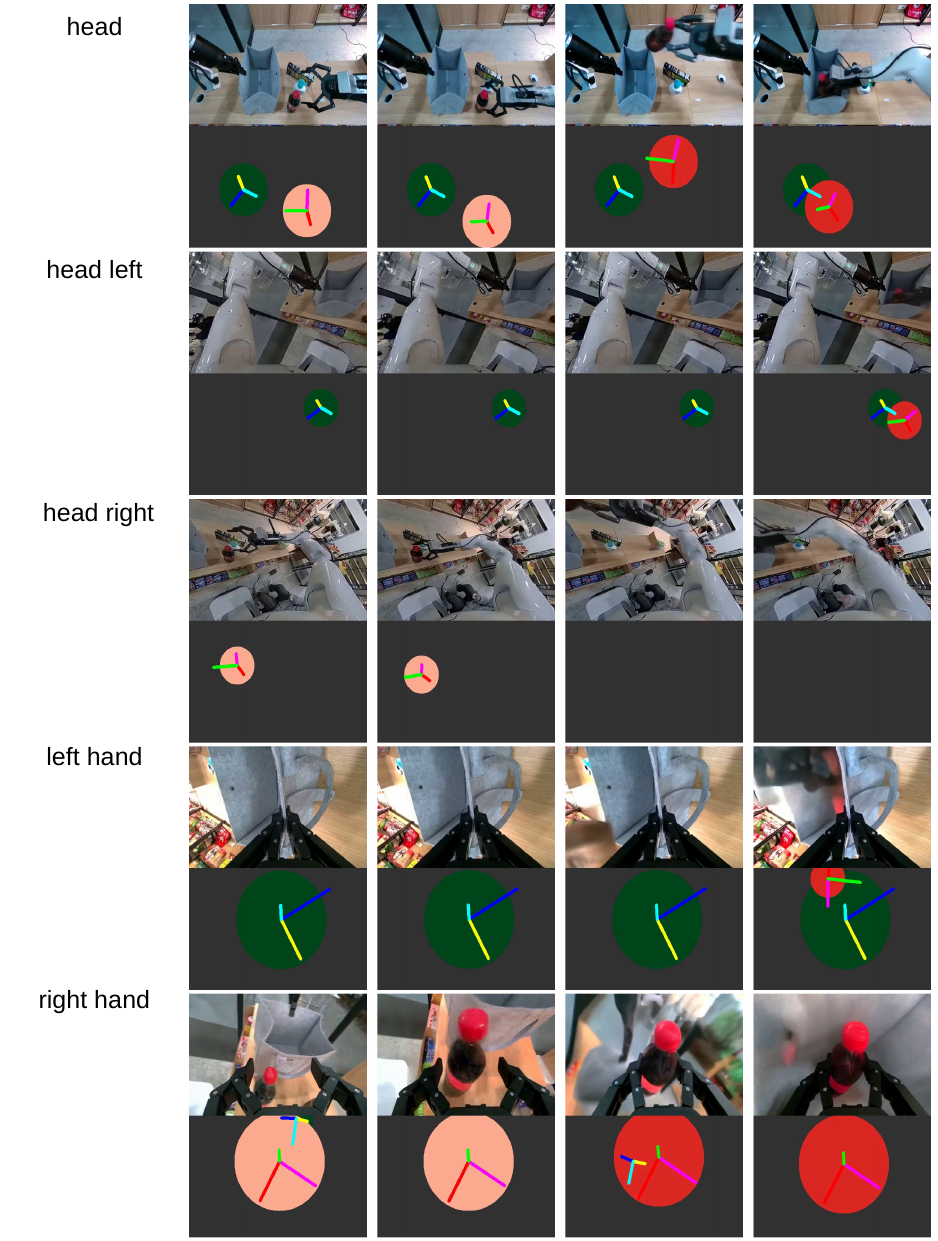}
    \caption{Multi-view generated videos. This task involves placing items from a desk into a bag, specifically packaging a Coke can. Each row displays each synchronized views generated by \Ours, showcasing consistent multi-perspective results at each timestep. We have shown 4 timesteps horizontally, illustrating the dynamic action sequence.}
    \label{fig:multi_view}
\end{figure}

In this section, we present an example of multi-view video generation, as shown in Figure~\ref{fig:multi_view}.

\clearpage

\subsubsection{Libero Results}

\begin{figure}[h] 
\centering
    \includegraphics[width=0.95\textwidth]{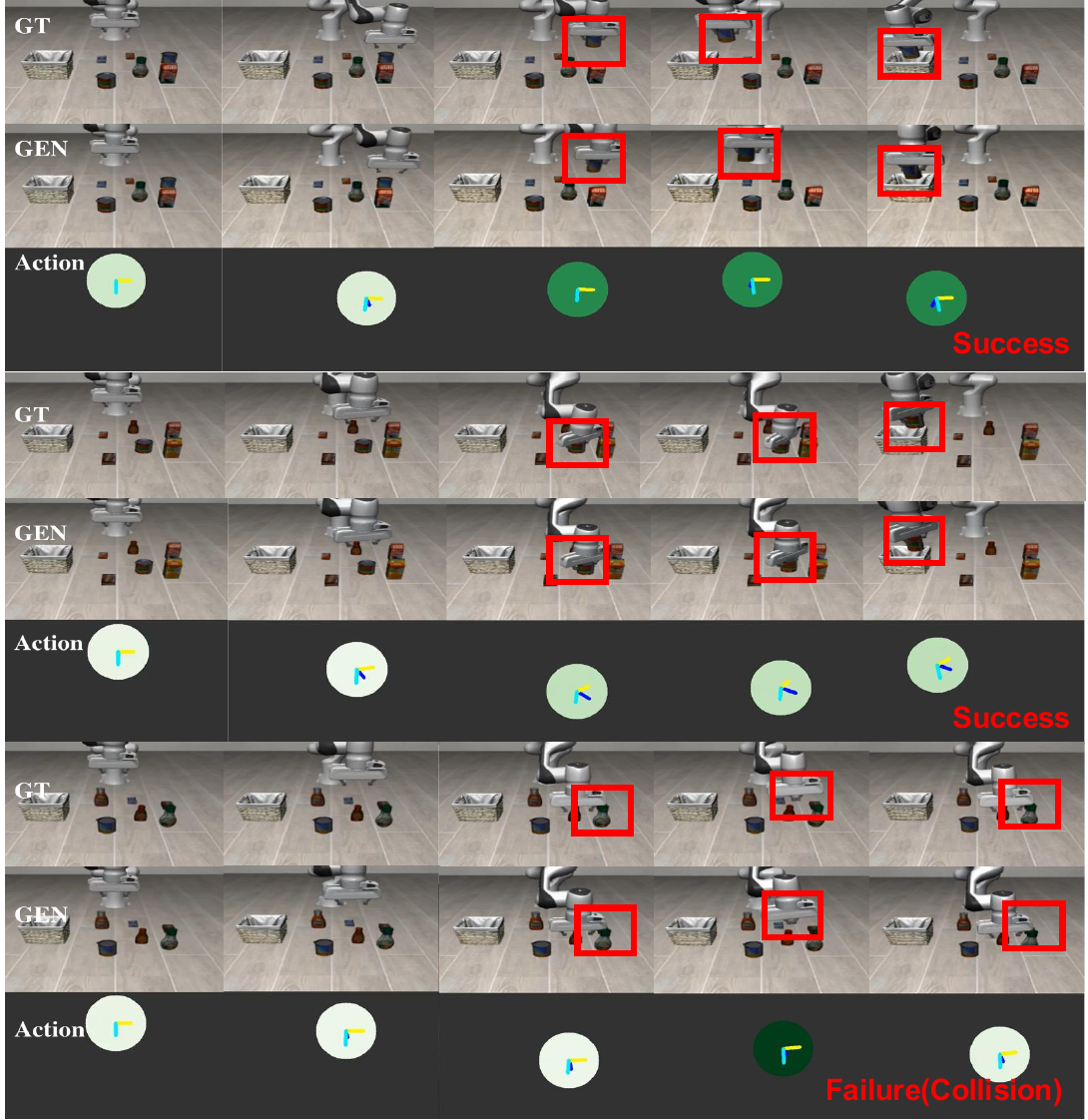}
    \caption{Comparison of results generated by Libero and \Ours with different action trajectories. Two successful cases of picking an item are shown, along with one failure case of an empty catch.}
    \label{fig:img_for_libero}
\end{figure}

To demonstrate the consistency and generality of \Ours, we fine-tuned the model using 417 trajectories and visualized generated action trajectories alongside Libero results. Qualitative results are shown in Figure~\ref{fig:img_for_libero}.

\clearpage

\subsubsection{Same Initial Conditions but Different Trajectories}

\begin{figure}[h]
    \includegraphics[width=1.\textwidth]{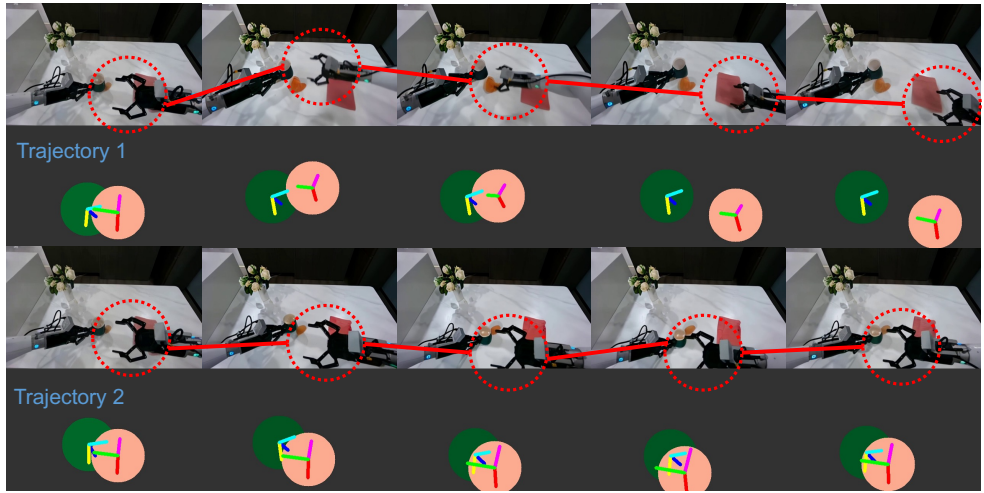}
    \vspace{-2cm}
    \caption{Results with the same initial conditions but different trajectories. The top row shows results for trajectory 1, while the bottom row shows results for trajectory 2.}
    \label{fig:img_for_multi_traj}
\end{figure}

To showcase the ability of precise trajectory-following, we present results generated under the same initial conditions but with different trajectories in Figure~\ref{fig:img_for_multi_traj}.

\clearpage

\clearpage

\subsection{Experiment Setup of \Ours's Application}
\label{sec:supp_application_setup}

\subsubsection{\Ours as a Policy Evaluator}

\begin{figure}[h!]
\centering
\includegraphics[width=1.\textwidth]{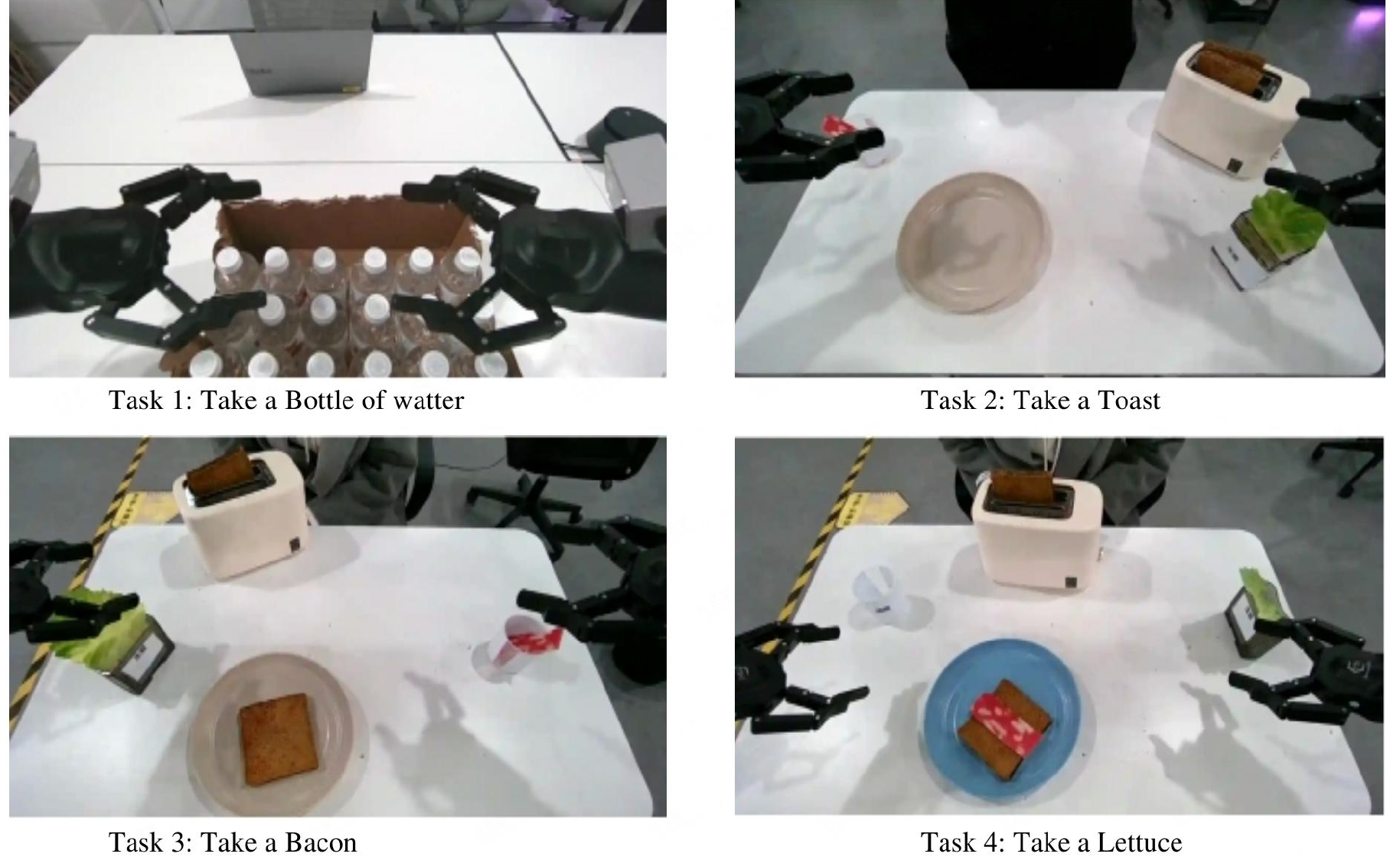}
\caption{Example of the initial conditions for the 4 tasks used in evaluation.}
\label{fig:img_for_eval_example}
\end{figure}

We train the official single-view Go-1 \cite{bu2025agibot} without the latent planner module and evaluate its performance in both real-world and simulated (\Ours) environments for comparison. During evaluation, the tester slightly randomizes the initial conditions of each task to obtain more generalized results. An example of the initial conditions for the 4 tasks is provided in Figure~\ref{fig:img_for_eval_example}. Each task is evaluated 40 times, with success defined as the robot successfully retrieving the target item.

\begin{figure}[h!]
\centering
\includegraphics[width=1.2\textwidth]{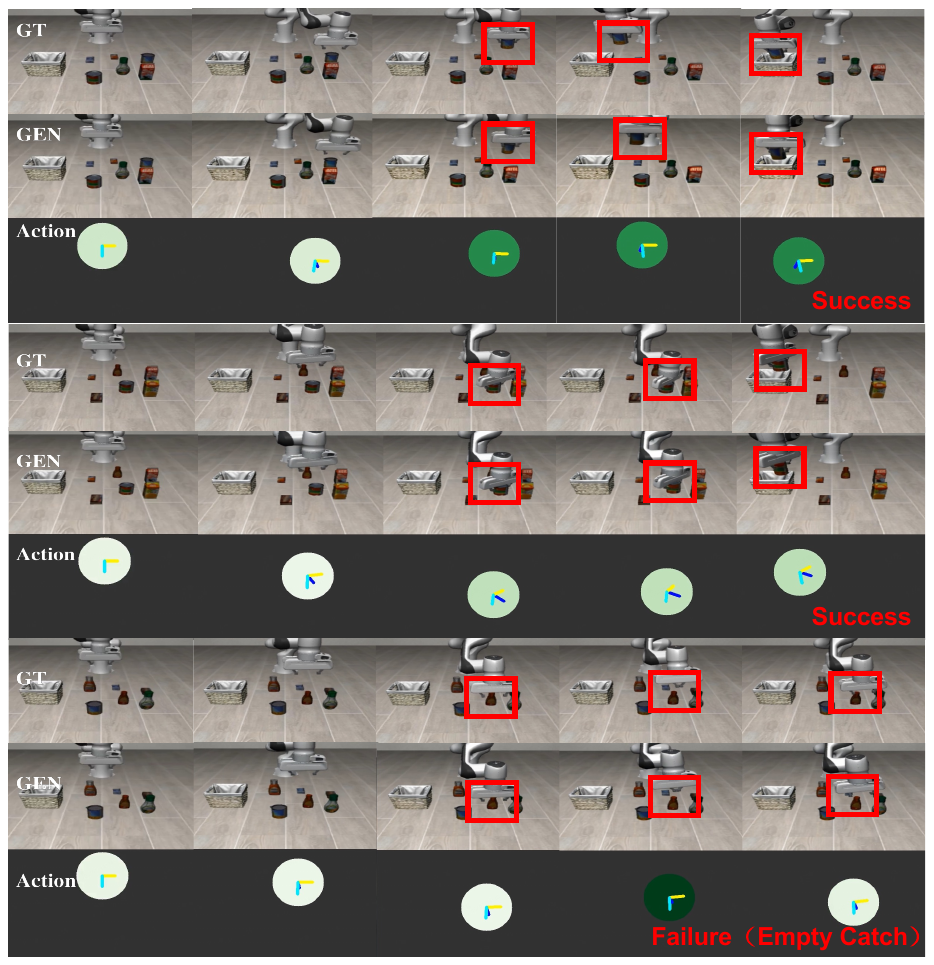}
\vspace{-4.5cm}
\caption{\textbf{Task 1}: Example of retrieving a bottle of water (failure case). The upper row shows the generated video from \Ours, and the lower row shows the rollout in real settings. Both are consistent in their results.}
\end{figure}

\begin{figure}[h!]
\centering
\includegraphics[width=1.2\textwidth]{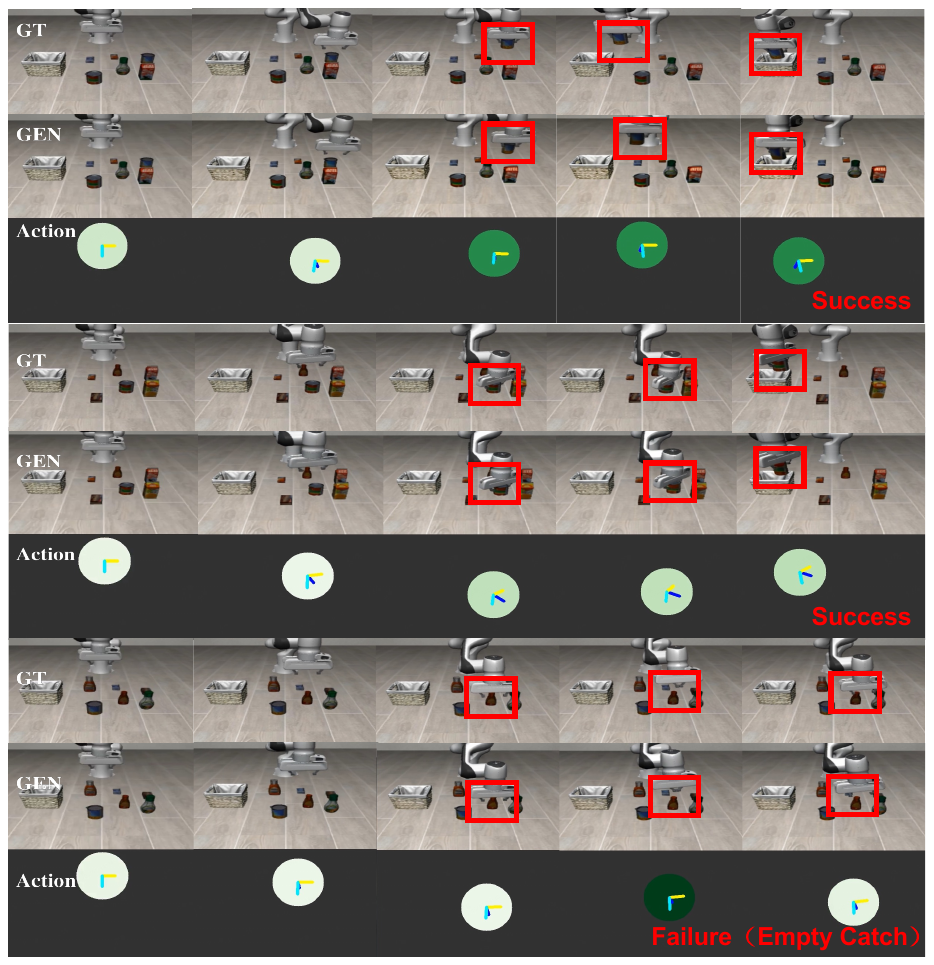}
\vspace{-4.5cm}
\caption{\textbf{Task 2}: Example of retrieving a piece of toast (success case). The upper row shows the generated video from \Ours, and the lower row shows the rollout in real settings. Both are consistent in their results.}
\end{figure}

\begin{figure}[h!]
\centering
\includegraphics[width=1.2\textwidth]{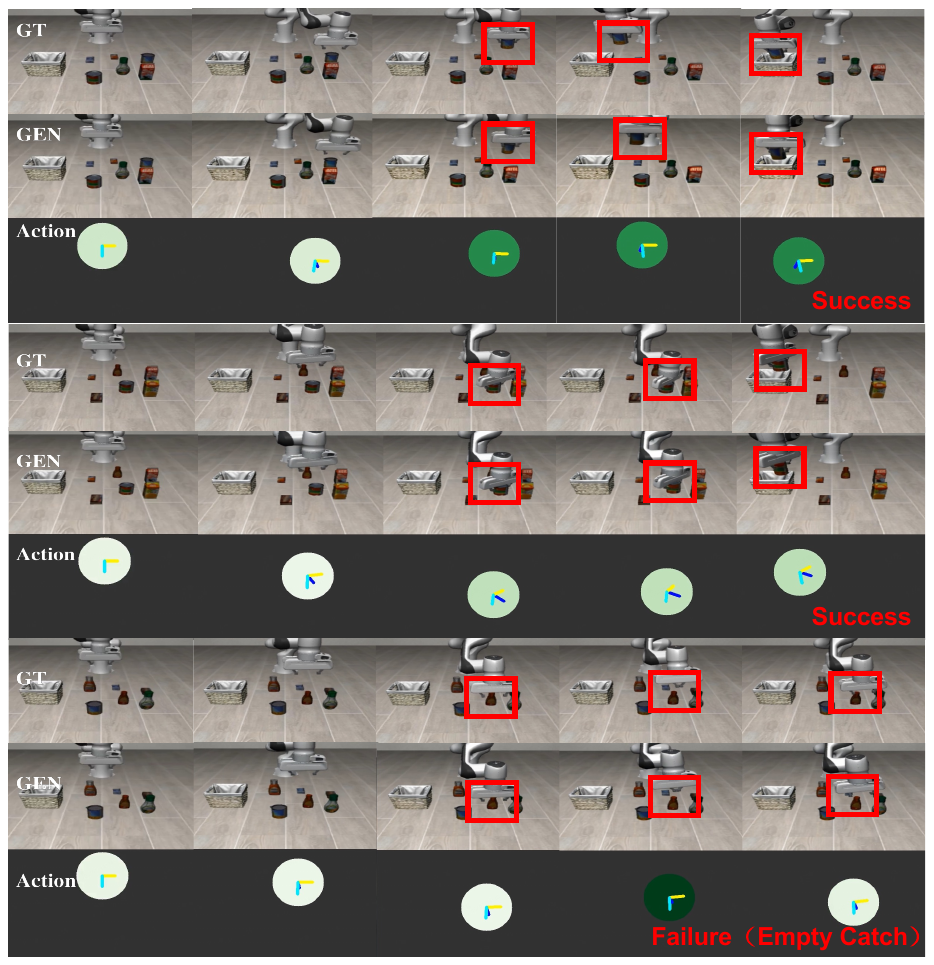}
\vspace{-4.5cm}
\caption{\textbf{Task 3}: Example of retrieving a ham slice (success case). The upper row shows the generated video from \Ours, and the lower row shows the rollout in real settings. Both are consistent in their results.}
\end{figure}
\clearpage
\begin{figure}[h!]
\centering
\includegraphics[width=1.2\textwidth]{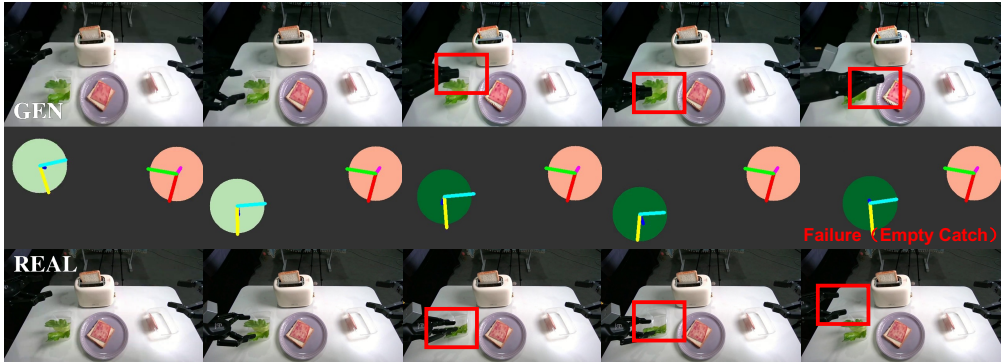}
\vspace{-4.5cm}
\caption{\textbf{Task 4}: Example of retrieving a lettuce leaf (failure case). The upper row shows the generated video from \Ours, and the lower row shows the rollout in real settings. Both are consistent in their results.}
\end{figure}

\subsubsection{\Ours as a Data Engine}

\begin{figure}[h]
\centering
\includegraphics[width=0.95\textwidth]{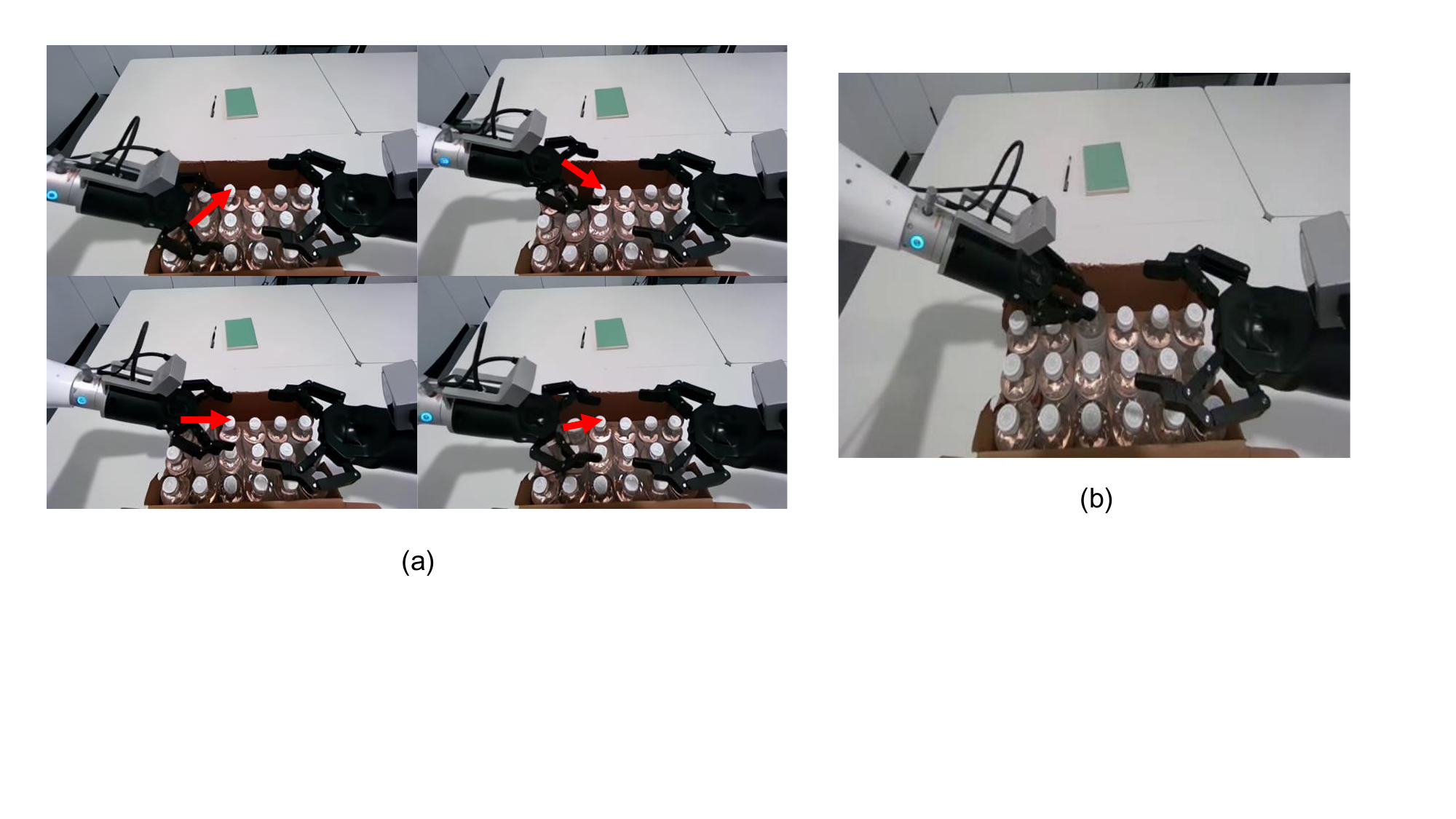}
\vspace{-2cm}
\caption{Example of the generated data for augmentation in Table~\ref{tab:data_augmentation}. Left: (a) Spatially augmented $a'_{t_{b-N}}$ with four frame examples. The red arrows indicate different directions from the synthetic image toward the target frame. Right: (b) Fixed $a_{t_b}$ frame. Frames between (a) and (b) are generated using linear interpolation.}
\label{fig:img_for_aug_example}
\end{figure}

In this section, we visually present examples of synthetic data used for data augmentation in training, as shown in Figure~\ref{fig:img_for_aug_example}. First, we randomly generate some  actions $a'_{t_{b-N}}$ within a fixed range from $a_{t_{b-N}}$. Then, using linear interpolation between $a'_{t_{b-N}}$ and $a_{t_b}$ to generate the action between these two. To generate the images, we first reverse the entire action sequence and use the real image as a condition for \Ours. This allows us to progressively generate images in reverse order. We then reverse the entire image sequence again for use by GO-1. Finally, we can get the image sequence from (Figure~\ref{fig:img_for_aug_example}a) to (Figure~\ref{fig:img_for_aug_example}b) to train a single-view Go-1 model.

\end{document}